\documentclass[11pt,a4paper]{article}

\usepackage[utf8]{inputenc}
\usepackage[T1]{fontenc}
\usepackage{lmodern}
\usepackage[margin=1in]{geometry}
\usepackage{amsmath,amssymb}
\usepackage{graphicx}
\usepackage{booktabs}
\usepackage{multirow}
\usepackage{enumitem}
\usepackage{xcolor}
\usepackage{caption}
\usepackage{subcaption}
\usepackage{tabularx}
\usepackage{array}
\usepackage{float}
\usepackage{fontawesome5}
\usepackage[most]{tcolorbox}
\usepackage[colorlinks=true,linkcolor=blue!60!black,citecolor=blue!60!black,urlcolor=blue!60!black]{hyperref}
\hypersetup{
  pdftitle={How Much Due Diligence Before You Bid? Learning in Intractable Takeover Auctions},
  pdfauthor={Zain Naboulsi},
  pdfsubject={Imperfect-information games, auctions, reinforcement learning},
  pdfkeywords={imperfect-information games, auctions, mergers and acquisitions, policy gradient, PPO, counterfactual regret minimization, exploitability, self-play, OpenSpiel}
}
\usepackage{natbib}

\definecolor{sparqcoral}{HTML}{C94A2A}
\definecolor{sparqcorallight}{HTML}{FFF5F2}
\definecolor{sparqcoralborder}{HTML}{E8634A}

\begin{document}

\begin{center}
  {\LARGE\bfseries\color{sparqcoral}%
    How Much Due Diligence Before You Bid?\\[0.3em]
    Learning in Intractable Takeover Auctions\par}
  \vspace{1em}
  {\large Zain Naboulsi}\\[0.3em]
  Principal AI Engineer, Sparq\\[0.2em]
  \texttt{zain.naboulsi@teamsparq.com}
\end{center}

\vspace{0.8em}

\begin{tcolorbox}[
  colback=sparqcorallight,
  colframe=sparqcoralborder,
  arc=3pt,
  boxrule=1pt,
  left=10pt, right=10pt, top=10pt, bottom=10pt,
  title={\textbf{Abstract}},
  fonttitle=\normalsize,
  coltitle=white,
  colbacktitle=sparqcoral,
  attach boxed title to top left={yshift=-2mm, xshift=5mm},
  boxed title style={arc=2pt, boxrule=0pt}
]
How much due diligence should a bidder buy before a takeover contest? Deal teams answer
this by judgment; we turn it into a computable, defensible number, cheap enough to run on
a commodity laptop. A bidder acts on private \emph{due diligence}: noisy, costly
estimates of a target whose true value no one observes. We study how to bid well under
that imperfect information, and what happens when the game grows past the reach of exact
solvers. The two are linked by one lever: each unit of diligence is another private
signal, and each signal multiplies the strategy space, so the same quantity that sets
how much diligence is worth buying also governs when the game outruns exact solvers. Our
answer is that the optimal amount is finite and falls as the per-signal cost rises,
including in an equilibrium where both bidders choose diligence, where competition makes
the smart amount more conservative still.
We cast takeover bidding as a small family of two-player zero-sum auction games from the
takeover-auction literature (common-value with a winner's curse, common-value with a
toehold, and independent-private-value), built on a reusable abstraction over OpenSpiel,
and benchmark nine solvers spanning exact, tabular, and deep families. Recent work argues
that simple generic policy-gradient methods rival specialized game-solving machinery
(counterfactual regret minimization, fictitious play, double oracle); we find that on
commodity CPU with no frontier-model spend, PPO and PPG are by far the strongest
\emph{learning} solvers, but only within that family: wherever the game is small enough
to tabulate, the exact solvers (CFR, MMD, PSRO) are both lower in exploitability and
faster. A scaling study shows the learning methods' flat per-target cost becomes an
advantage only beyond exact enumeration. Solving the auction's genuine own-profit
Bayes-Nash equilibrium exactly, we then read off the value of a bidder's diligence and
the cost at which acquiring more of it stops paying. Finally, on a multi-signal auction
too large to enumerate, where exact solvers cannot run, PPO and PPG drive a calibrated
learned-best-response exploitability estimate to its resolution floor, below a naive
unshaded bidder; we report it as a lower bound, not a Nash certificate. We release the
games, solvers, and experiments.

\vspace{0.8em}
\noindent\begin{minipage}[b]{0.82\linewidth}
  \footnotesize\raggedright\mbox{\faGithub\ \url{https://github.com/zainnab-sparq/imperfect-information-deal-games}}
\end{minipage}%
\hfill
\begin{minipage}[b]{0.16\linewidth}
  \raggedleft
  \includegraphics[height=1cm]{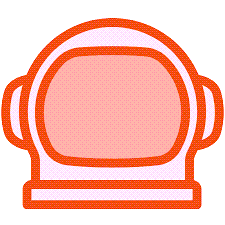}\\[-2pt]
  {\small\textsf{\textbf{Sparq}}}
\end{minipage}
\end{tcolorbox}

\vspace{0.4em}

\section{Introduction}

A merger or acquisition is rarely a posted price. It is a contested process in
which an acquirer and a target, or two competing acquirers, hold private
information and act strategically under uncertainty about each other and about the
asset. Economists model these as games of incomplete information \citep{harsanyi1967}:
takeover contests as auctions with private or common values, negotiations as bargaining
under asymmetric information. The computational game-solving community has produced
powerful algorithms for large imperfect-information games, but has concentrated on
recreational benchmarks rather than economic ones.

The question in our title is a real one for a deal team, and it carries a hidden
computational edge. Each unit of due diligence a bidder performs is one more private
signal it brings to the auction, and each signal multiplies the number of distinct
situations it must plan for, so the strategy space grows exponentially in the amount of
diligence. The economic question, how much diligence is worth buying, and the
computational one, when the game stops being exactly solvable, are therefore governed by
the same lever: the number of signals a bidder carries. They are not the same number,
though; we find the economically optimal amount of diligence is modest, while the game
outruns exact solvers only at much larger signal counts, so the two questions meet on one
axis but at different points along it. Deal teams answer the diligence question by judgment.
This paper answers the economic side directly, solving the auction's own-profit
equilibrium to find that the optimal amount is finite and falls with the per-signal cost,
which turns that judgment call into a defensible, model-backed number and points to where
a firm that keeps researching is spending past the point it pays. It does so cheaply, on
a commodity laptop with no frontier-model spend, and uses the computational side to
stress-test cheap generic solvers right at the boundary where exact methods give out.

Two facts motivate this paper. First, the standard specialized solvers
(counterfactual regret minimization, fictitious play, double oracle, and their deep
descendants) are expensive and delicate to deploy. Second, recent work
\citep{rudolph2026} argues that in two-player zero-sum imperfect-information games,
\emph{simple generic policy-gradient methods} such as PPO are competitive with or
superior to that machinery. If so, modeling deal-making under imperfect information
need not require frontier models or elaborate solvers; a cheap self-play loop on
commodity hardware may suffice.

We test this in the deal-making setting. Our contributions are: (1)~a small family
of zero-sum deal games grounded in the takeover-auction literature, sharing a
reusable \texttt{DealGame} abstraction; (2)~a like-for-like benchmark of nine
solvers under exact exploitability, on iterations \emph{and} wall-clock, spanning
the exact (CFR, MMD, PSRO), tabular-learning (REINFORCE), and deep-learning (PPO,
PPG, a generic deep policy gradient, Deep CFR, NFSP) families; (3)~evidence that the
generic policy-gradient methods PPO and PPG are the strongest learning solvers,
dominating deep CFR and deep fictitious play at equal episode budgets, while the
exact solvers remain best wherever the game is small enough to tabulate; (4)~a
scaling study isolating why the learning methods are nonetheless of interest (their
per-target wall-clock is roughly constant in game size where exact CFR's grows
steeply), with the honest caveat that within enumerable sizes PPO never overtakes CFR
and fails to reach the target on the largest; together with a first step into the
genuinely intractable regime, a multi-signal auction too large to enumerate, where PPO
and PPG drive a \emph{calibrated} learned-best-response exploitability estimate to its
resolution floor, below a naive unshaded bidder, in a regime exact solvers cannot
enter; and (5)~economic results read off the auction's genuine own-profit Bayes-Nash
equilibrium, which we solve exactly rather than through the zero-sum rendering: the
value of a bidder's diligence is positive on net, and, charging a per-signal cost, the
profit-maximizing amount of due diligence is finite and falls as cost rises across every
parameterization we check, including when both bidders choose diligence in a symmetric
acquisition equilibrium (which competition makes more conservative than a one-sided
calculation), giving a model-internal answer to the question in our title (its fine
schedule is game-specific and sensitive to the bid-grid resolution); the toehold's
robust effect at the true equilibrium is on value rather than on equilibrium bidding. Everything runs on
a commodity laptop CPU, with no frontier-model API calls: individual experiments take
minutes to tens of minutes, the full benchmark suite a few hours, and the economic
equilibria solve in seconds.

Relative to \citet{rudolph2026}, who establish the policy-gradient thesis on
recreational benchmarks at tabulatable scale, our contribution is to port it to
economic deal games, to tie the exact-versus-learning crossover to tree-enumerability
through a wall-clock scaling law, and to take a first step past enumeration with a
learned-best-response lower bound that we calibrate against exact exploitability.

\section{Related work}

\paragraph{Solving imperfect-information games.} Counterfactual regret minimization
\citep{cfr2007} is the tabular standard; Monte Carlo CFR \citep{mccfr2009} samples
it, predictive (optimistic) variants accelerate it by extrapolating the next gradient
\citep{predictivecfr2021}, and Deep CFR \citep{deepcfr2019} scales it with neural
advantage and strategy networks. ESCHER \citep{escher2022} and DREAM \citep{dream2020}
are later variance-reduced deep-CFR-family methods. Neural Fictitious Self-Play (NFSP)
\citep{nfsp2016} and Policy-Space Response Oracles (PSRO) \citep{psro2017} bring
fictitious play and double oracle to the deep setting; CFR-family search underlies the
superhuman poker agent Libratus \citep{libratus2018}, and ReBeL \citep{rebel2020}
combines RL with search. Magnetic mirror descent (MMD) \citep{mmd2023} is a regularized
method competitive as both a tabular and a deep solver; the same regularization idea,
scaled with deep RL as regularized Nash dynamics, masters Stratego model-free and
without tabulation \citep{deepnash2022}. PPO (Proximal Policy
Optimization) \citep{ppo2017} and its phasic variant PPG (Phasic Policy Gradient)
\citep{ppg2021} are the generic policy-gradient methods that
\citet{rudolph2026} argue are competitive with all of these; we evaluate both. We
build on OpenSpiel \citep{openspiel2019}.

\paragraph{Auctions and takeovers.} Auction theory characterizes bidding under
private and common values \citep{klemperer1999}, from the common-value model of
\citet{wilson1977} to the affiliated-values generalization of \citet{milgromweber1982},
which is closest to the game we build. Endogenous information acquisition before bidding,
the formal version of our diligence question, is studied by \citet{persico2000} and,
in the broader mechanism-design and auction-format literature, by
\citet{bergemann2002} and \citet{comptejehiel2007}; we take
signal precision as exogenous and add a per-signal cost only in the diligence-cutoff
study. The takeover literature studies \emph{sequential} preemptive and jump bidding
\citep{fishman1988} and toehold effects \citep{bulow1999}; our sealed simultaneous auction
deliberately abstracts away the sequential signaling that drives preemption, and
isolates the toehold's payment-side channel. Bilateral trade under private
information is bounded by impossibility results \citep{myerson1983}. A line of work
learns equilibria of auction games directly, with neural pseudogradient ascent
\citep{npga2021} and first-order gradient dynamics \citep{kohring2023}, and learning
has likewise been applied to negotiation and
multi-issue bargaining \citep{bargaining2020}. We borrow these structures as game
definitions rather than deriving closed-form equilibria.

\section{Deal games}

\paragraph{Reusable abstraction.} A \texttt{DealGame} layer centralizes the two
pieces identical across deal games and easy to get wrong: information-set
construction (a player's information state is built from its own private tokens plus
public tokens, so an opponent's private draw never leaks in, which would silently
invalidate exploitability), and the zero-sum payoff contract. Concrete games supply
only their move protocol and economics.

\paragraph{Common-value takeover auction.} Two bidders compete for a target of
common value $W$, unknown to both; each receives a private signal of $W$ with
per-bidder noise (the affiliated common-value setting of \citealp{milgromweber1982}).
Sealed first-price bids are submitted simultaneously; the higher bid wins, pays its
bid, and earns $W-\text{bid}$; the loser earns nothing. The common-value structure
induces a winner's curse \citep{capen1971, kagellevin1986}: conditioning on winning
is bad news about $W$, so a bidder that does not shade overpays in expectation.

\paragraph{A note on the zero-sum rendering.} We score each game zero-sum as the
\emph{difference} in profit, the head-to-head benchmark of \citet{rudolph2026}, so
that exact two-player exploitability applies. This is a deliberate modeling choice
with a cost: unlike an equal-split of a fixed pie, subtracting the opponent's
(endogenous) profit is not an incentive-preserving affine shift; it adds a rivalry
term, so the equilibrium we compute is the Nash equilibrium of the relativized game,
not the Bayes-Nash equilibrium of the underlying general-sum auction. We use it for
methodological reach in the solver benchmark, where exact two-player exploitability is
the metric; but where an economic quantity is at stake (the information-asymmetry,
diligence, and toehold studies, \S\ref{sec:experiments}) we read the result off the
auction's genuine own-profit Bayes-Nash equilibrium, which we solve exactly
(\S\ref{sec:methods-gs}), not the relativized one. The \texttt{DealGame} abstraction
supports the general-sum contract directly.

\paragraph{Toehold variant.} A bidder holding a toehold $\theta$ buys only the
remaining $(1-\theta)$ if it wins and collects $\theta$ of the price if it loses;
$\theta=0$ recovers the base auction.

\paragraph{Private-value auction.} A structurally different game: each bidder draws
and observes its own independent private value, with no common value and no winner's
curse. Including it guards against artifacts of the common-value structure.

\section{Methods}

We benchmark nine solvers, grouped into three families. All are evaluated by the
same exact metric: OpenSpiel exploitability, i.e. NashConv halved, the average over
the two players of how much each could gain by best-responding to the other. It is
zero at a Nash equilibrium and lower is better; we compute it by exact tree
traversal.

\paragraph{Exact and tabular.} CFR \citep{cfr2007}; magnetic mirror descent (MMD)
\citep{mmd2023} as a Nash solver; PSRO \citep{psro2017} with exact best-response
oracles and a projected-replicator-dynamics meta-solver (exact oracles are the
strongest, and appropriate for games small enough to tabulate); and a from-scratch
tabular policy gradient (REINFORCE with a per-information-state baseline and entropy bonus).

\paragraph{Deep (function approximation).} PPO \citep{ppo2017}, the seed paper's
method, implemented as clipped self-play with a value baseline, multiple epochs over
minibatches, and entropy regularization; PPG \citep{ppg2021}, its phasic variant,
which decouples policy and value optimization with periodic auxiliary value-distillation
phases under a behavioral-cloning constraint; a generic deep policy gradient
(OpenSpiel's PolicyGradient agent with the regret-policy-gradient loss, ``DeepPG''
in tables), the deep analogue of our tabular REINFORCE; Deep CFR \citep{deepcfr2019}
(``DeepCFR'' in tables), the deep counterfactual-regret baseline standing in for the
deep-CFR family (including ESCHER \citealp{escher2022} and DREAM \citealp{dream2020});
and NFSP \citep{nfsp2016}, the fictitious-play baseline. Because each bidder takes exactly one
action per episode (its bid), the policy-gradient advantage reduces to the Monte-Carlo
return minus the value baseline, with no multi-step bootstrapping; the clipped surrogate,
minibatch epochs, and entropy bonus are otherwise standard.

\paragraph{Tuning and fairness.} We selected the learning rate of the
policy-gradient family (PPO, PPG, DeepPG) by a coarse sweep, because the generic
policy gradient is sensitive to it (see \S\ref{sec:experiments}); the other deep
baselines (Deep CFR, NFSP) use author-recommended settings adapted to the short CPU
budget rather than a per-game sweep. We report this asymmetry plainly so the
comparison is not misread as exhaustively tuned on all sides. The policy-gradient
methods (PPO, PPG, DeepPG) and NFSP share the same 64-unit single-hidden-layer scale;
Deep CFR uses its standard two-layer $(64, 64)$ advantage and policy networks. All
methods are evaluated identically.

\paragraph{Iterate averaging.} In two-player zero-sum games the last iterate of
policy gradient cycles around the equilibrium and does not converge in
exploitability; only the time-average converges. We therefore evaluate the deep
policy-gradient methods (PPO, PPG, DeepPG) on a tabular tail-average of the network's
per-information-state action probabilities (the games are small enough to tabulate a
neural policy exactly), reporting that average as the convergent metric. The other
deep baselines are evaluated on their own native convergent objects for the same
reason: NFSP on its average-policy network and Deep CFR on its average-strategy
network. The tail-average discards the first half of training (a burn-in) so the
random-init transient does not pollute it; we note that this gives the
policy-gradient family a head start the always-on averages of NFSP/Deep CFR do not
get, and report both the tail-average and the oscillating last iterate. To confirm the
headline ordering is not an artifact of that head start, we also score the
policy-gradient family with \emph{no} burn-in (a full average that includes the
random-init transient, the same treatment the NFSP/Deep CFR averages receive): on the
headline game PPG and PPO then read $0.011$ and $0.022$ (three seeds), essentially
unchanged from their tail-averages and still far below NFSP ($0.507$) and Deep CFR
($0.121$), so the gap is a method effect rather than a burn-in advantage. Deep methods
are stochastic, so we report mean $\pm$ standard deviation over five seeds; we record
wall-clock seconds alongside episodes for a compute-normalized comparison.

In the intractable regime the policy cannot be tabulated, so there we tail-average the
network \emph{weights} instead. Averaging the weights of a nonlinear network is not the
same object as the time-average of the induced policy, and the convergence argument
above is about the latter, so we validate weight-averaging on tractable CV-large, where
exact exploitability is available: from a single PPO trajectory we average the same
post-burn-in iterates both ways. The weight tail-average reaches exact exploitability
$0.026 \pm 0.006$, close to the tabular time-average's $0.021 \pm 0.009$ and far below
the cycling last iterate's $0.052 \pm 0.026$ (three seeds). Weight-averaging is thus a
sound stand-in for the convergent time-average where the latter cannot be built.

\paragraph{General-sum equilibrium.}
\label{sec:methods-gs}
The economic comparative statics are properties of the underlying \emph{general-sum}
auction, not of the zero-sum rendering the solver benchmark uses, so for them we solve
the auction's own-profit Bayes-Nash equilibrium exactly. The auction factorizes:
terminal profit depends only on the common value and the two bids, and each bidder's
signals are conditionally independent given the value, so the game collapses to small
dense tensors (one row per information set) and the equilibrium is found by fictitious
play on each bidder's \emph{own} profit, with no tree traversal and no Monte-Carlo
noise. We report the achieved NashConv (the summed own-profit gain from a unilateral
deviation; below $\sim$$10^{-2}$ throughout, and $\sim$$10^{-4}$ except at the
fully-informed corner) so convergence is measured rather than assumed, and pin the
tensor model to the OpenSpiel game with a unit test that matches expected own profit on
several game shapes. This lets each economic claim be read off the genuine equilibrium
rather than a one-sided deviation or the relativized objective; because the tensors are
tiny it also runs in a fraction of a second, including the per-bidder asymmetric-signal
configurations the OpenSpiel game cannot express.

\section{Experiments}
\label{sec:experiments}

All experiments run on CPU in a Docker image that pins the Python base image,
OpenSpiel, and PyTorch versions and fixes the CPU thread count
(\texttt{OMP\_NUM\_THREADS} and \texttt{torch.set\_num\_threads}) so multithreaded
float reductions are bounded; the residual nondeterminism that remains is why we still
report over seeds. Full hyperparameters are in Appendix~\ref{app:hparams}. The
deep self-play, cross-game, zero-sum toehold, and scaling studies are driven by one
benchmark script; the tabular convergence figure comes from a small auxiliary script;
and the general-sum economic results (information asymmetry, diligence, and the toehold
at the own-profit equilibrium) come from the exact equilibrium solver of
\S\ref{sec:methods-gs}.

\begin{figure}[t]
\centering
\includegraphics[width=0.6\linewidth]{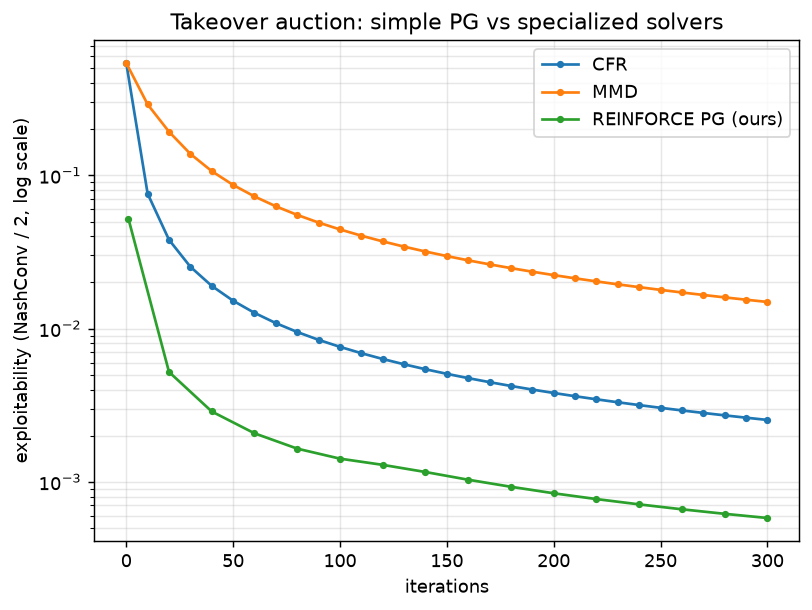}
\caption{Tabular convergence on the small common-value auction: exploitability
(NashConv halved, log scale, lower is better) vs.\ iterations for CFR, MMD, and our
from-scratch REINFORCE policy gradient. From the auxiliary experiment script.}
\label{fig:conv}
\end{figure}

\paragraph{Tabular convergence.} On the small common-value auction (from the
auxiliary script) all exact/tabular methods drive exploitability toward zero
(Fig.~\ref{fig:conv}). Final exploitability: CFR \textbf{0.0025}, MMD \textbf{0.015},
tabular REINFORCE \textbf{0.0006}. On this small game the from-scratch policy
gradient is competitive with the specialized solvers; the cross-game table below
shows it degrades on larger games at a fixed budget.

\begin{figure}[t]
\centering
\includegraphics[width=0.95\linewidth]{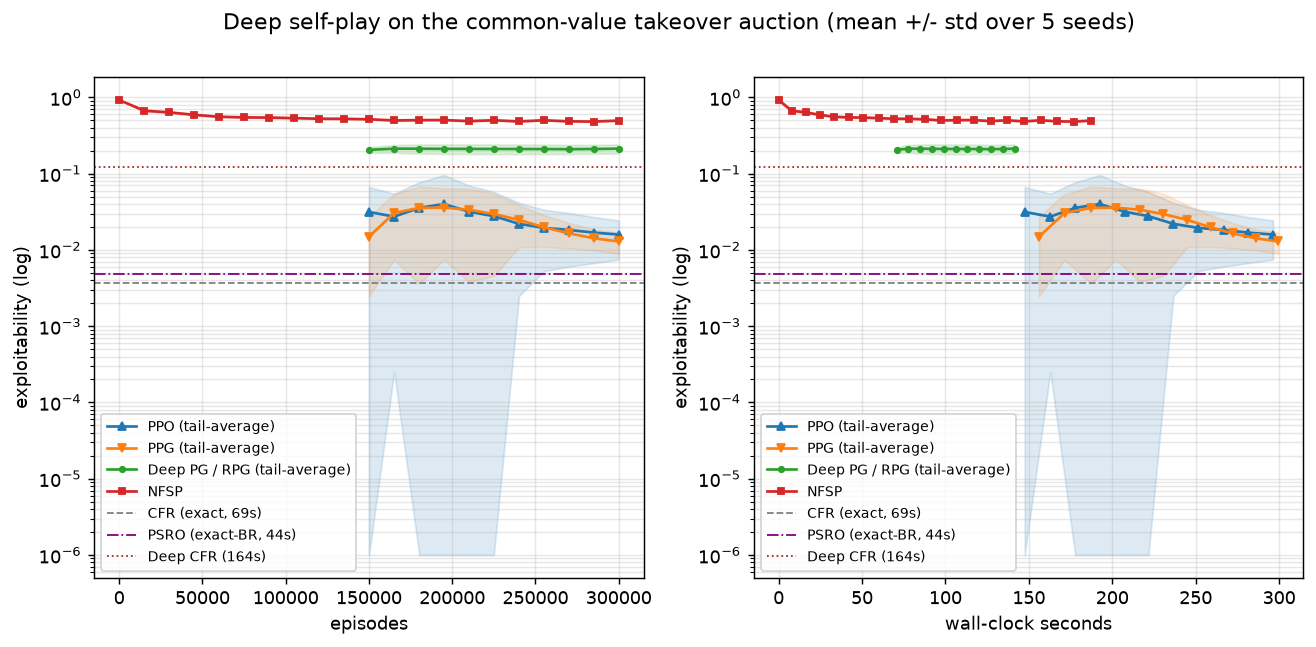}
\caption{Deep self-play on the common-value auction (5 values, 6 bids):
tail-averaged exploitability (log scale, lower is better) vs.\ episodes (left) and
training wall-clock seconds (right), mean $\pm$ std over five seeds, for PPO, PPG,
DeepPG, and NFSP, with exact CFR/PSRO and Deep CFR shown as horizontal references
(their wall-clock in the legend).}
\label{fig:deep}
\end{figure}

\paragraph{Deep self-play (headline).} On a larger common-value auction (5 values,
6 bid levels), over five seeds, the two generic policy-gradient methods are the
strongest learners: PPG converges to tail-averaged exploitability
\textbf{$0.013 \pm 0.004$} and PPO to \textbf{$0.016 \pm 0.008$}, an order of
magnitude below the generic deep policy gradient's \textbf{$0.213 \pm 0.029$},
roughly $8\times$ below Deep CFR's \textbf{$0.122 \pm 0.007$}, and about $30\times$
below NFSP's \textbf{$0.496 \pm 0.004$} (Fig.~\ref{fig:deep}). They are also markedly
more sample-efficient, reaching their plateau in far fewer episodes. The exact
references remain both lower and faster: CFR \textbf{0.0037} in \textbf{69}s and PSRO
\textbf{$0.0048 \pm 0.0004$} in \textbf{44}s, against roughly $300$s for PPO/PPG.
Indeed the policy-gradient methods are the most expensive \emph{per episode} (PPO/PPG
$\approx 300$s for $300$k episodes versus DeepPG $141$s and NFSP $187$s), so their
edge over the deep baselines is at equal \emph{episode} budgets, not equal
wall-clock; on a game this small the exact solvers dominate on both axes. A practical
caveat behind the generic policy gradient: with an aggressive learning rate its last
iterate diverges into a limit cycle near exploitability $0.40$; a smaller learning
rate (which we use) restores monotone convergence, and the tail-average (the
convergent quantity in zero-sum games) gives the stable, low-variance numbers
reported here.

\paragraph{Across games.} Table~\ref{tab:cross} reports final exploitability for
every method on every game. Two patterns hold. Among the learning methods, PPO and
PPG are the strongest and trade places (PPG lowest on CV-small and CV-large, PPO on
CV-toehold and PV), and both sit far below the generic deep policy gradient, Deep
CFR, and NFSP on all four games. The gaps are large and consistent across five
seeds, so we read these as robust orderings rather than precise estimates. As
expected for games small enough to tabulate, the exact solvers (CFR, MMD, PSRO)
achieve the lowest exploitability overall; tabular REINFORCE is competitive only on
the smallest game and degrades sharply on the larger ones. The case for the learning
methods is therefore not that they win here, but that they carry over unchanged as
games grow (\S\ref{sec:scaling}).

\begin{figure}[t]
\centering
\includegraphics[width=0.62\linewidth]{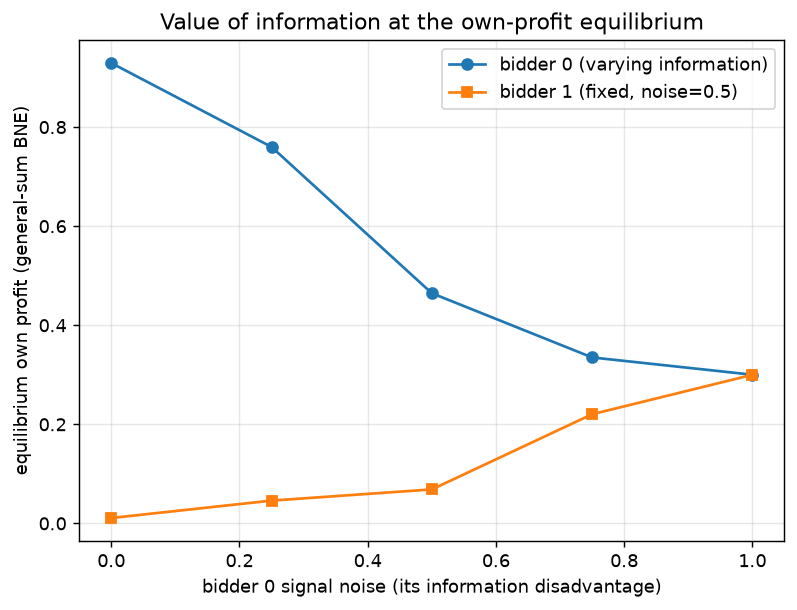}
\caption{Value of information at the genuine own-profit Bayes-Nash equilibrium (solved
exactly, \S\ref{sec:methods-gs}), not the zero-sum rendering. As bidder~0's signal
noise rises (bidder~1 fixed at noise $0.5$), bidder~0's equilibrium own profit falls
and the better-informed rival's rises, meeting where both are uninformed. The value of
information is positive, steepest at intermediate signal quality, and the uninformed
bidder still captures real surplus.}
\label{fig:asym}
\end{figure}

\paragraph{Information asymmetry.} We read the value of information off the auction's
genuine own-profit Bayes-Nash equilibrium (\S\ref{sec:methods-gs}), not the zero-sum
rendering or a one-sided deviation. Holding bidder~1's signal noise at $0.5$ and
sweeping bidder~0's, bidder~0's equilibrium own profit falls monotonically as its
signal degrades, from \textbf{0.93} with a perfect signal (the one corner the solver
brings only to a near-equilibrium, NashConv $\sim$$8\times10^{-3}$; \S\ref{sec:methods-gs})
to \textbf{0.30} with a
useless one, while the now-better-informed rival's value rises to meet it; at equal
(useless) information the two split the surplus (\textbf{0.30} each, Fig.~\ref{fig:asym}).
This is the value of information measured in a real equilibrium: positive, steepest at
intermediate signal quality, and economically meaningful even at the floor, because a
poorly informed bidder still captures real surplus. The zero-sum rendering and a one-sided own-profit best response
agree on the direction of the decline (the zero-sum value falls from \textbf{+0.389}
to \textbf{+0.002}, the one-sided best response from \textbf{+0.389} to \textbf{+0.168}),
but both overstate the collapse: the rivalry term and the fixed-opponent assumption each
understate the residual surplus the uninformed bidder retains in the true equilibrium.
This is exactly the gap the exact general-sum solver closes.

\begin{figure}[t]
\centering
\includegraphics[width=0.95\linewidth]{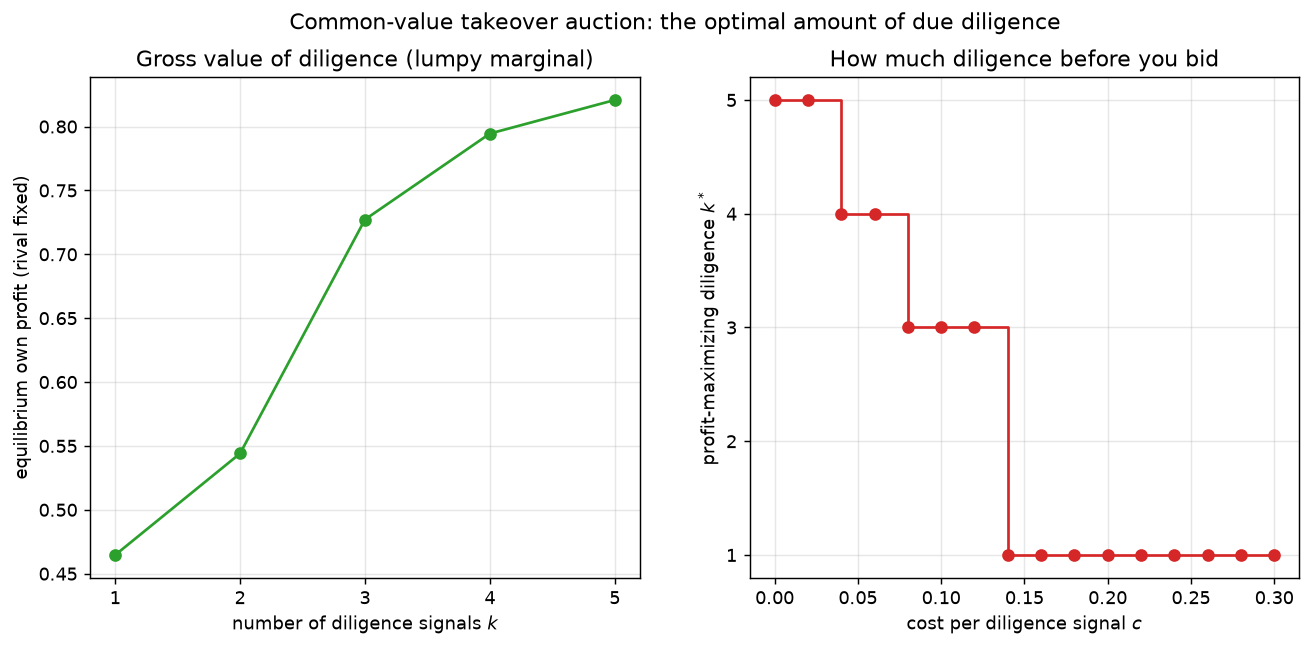}
\caption{How much due diligence, in the base game (five values, six bids, noise $0.5$).
Left: bidder~0's equilibrium own profit as it acquires more independent diligence
signals $k$ (rival fixed at one signal), rising with a lumpy marginal. Right: with a
per-signal diligence cost $c$, the profit-maximizing amount of diligence
$k^\star=\arg\max_k[\text{value}(k)-c\,k]$, a step function falling from $5$ to $1$ as
cost rises. The cost-benefit cutoff the title asks for; its robustness across
parameterizations is Fig.~\ref{fig:dilig-robust}.}
\label{fig:diligence}
\end{figure}

\paragraph{How much due diligence?} The value of information is what a bidder pays
\emph{for}; the title's question is how much to buy. We model diligence as the number
of independent signals $k$ a bidder acquires before bidding. At the own-profit
equilibrium (rival fixed at one signal), in our base game (five values, six bids,
signal noise $0.5$) bidder~0's equilibrium profit rises with its own $k$, from
\textbf{0.46} at $k=1$ to \textbf{0.82} at $k=5$ (Fig.~\ref{fig:diligence}, left), with
a \emph{lumpy} marginal: here the third signal is worth more than the second ($+0.18$
versus $+0.08$). Charging a per-signal diligence cost $c$, the profit-maximizing amount
of diligence is $k^\star=\arg\max_k\,[\text{value}(k)-c\,k]$ (Fig.~\ref{fig:diligence},
right): in this game it steps down from $k^\star=5$ when signals are nearly free to
$k^\star=1$ once $c$ exceeds about $0.14$ per signal, skipping $k=2$ entirely. This is
the cost-benefit cutoff the title asks for. The load-bearing claim is not merely that
the optimum is finite (with a strictly positive per-signal cost and a bounded marginal
value of information, finiteness is close to automatic) but that it \emph{falls
monotonically} as the cost rises; the specific schedule (which $k$ are skipped) is the
fragile part, as the next paragraphs show.

\begin{figure}[t]
\centering
\includegraphics[width=0.95\linewidth]{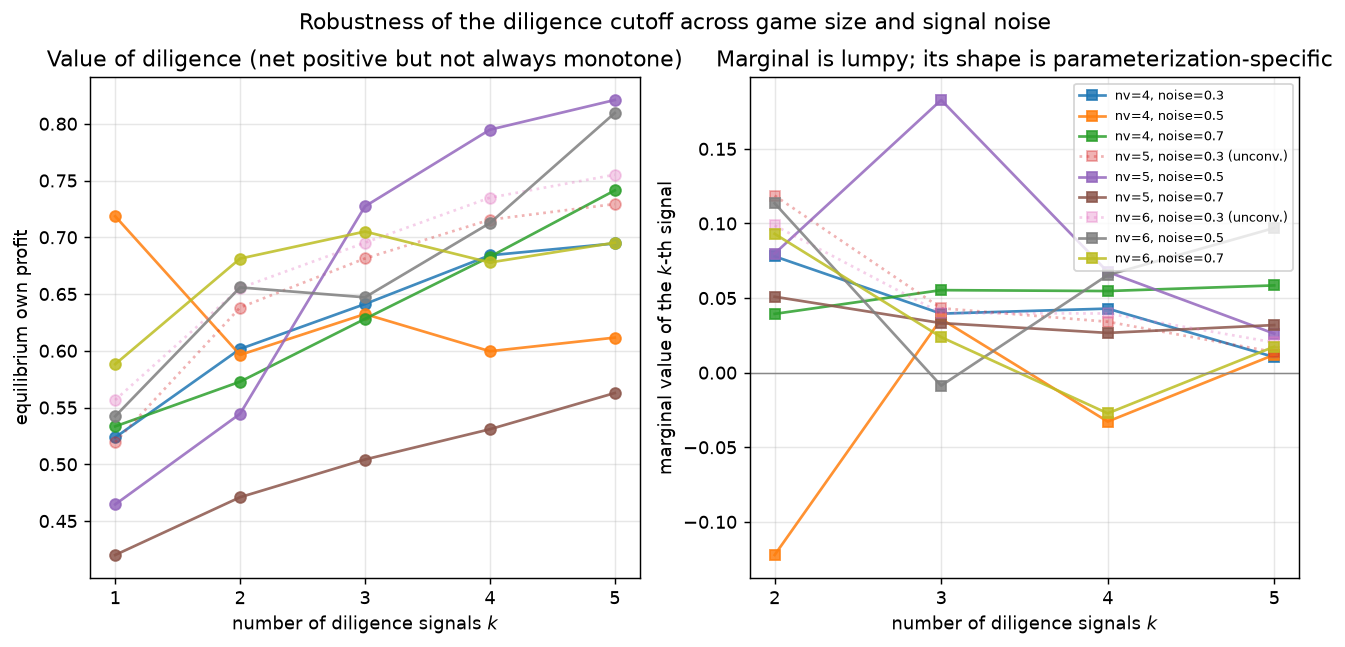}
\caption{Robustness of the diligence cutoff across game sizes (four, five, six common
values) and signal qualities (noise $0.3$, $0.5$, $0.7$); dotted lines are
parameterizations the fictitious-play solver did not bring to tolerance and are excluded
from the counts. Left: the value of diligence is positive on net but not always monotone
in $k$. Right: the marginal value is lumpy, but its shape (which signals are worth most)
is parameterization-specific. The cutoff itself is finite and falls with cost in every
converged case ($7/7$); the specific $k$ it skips is not a general feature.}
\label{fig:dilig-robust}
\end{figure}

\begin{figure}[t]
\centering
\includegraphics[width=0.6\linewidth]{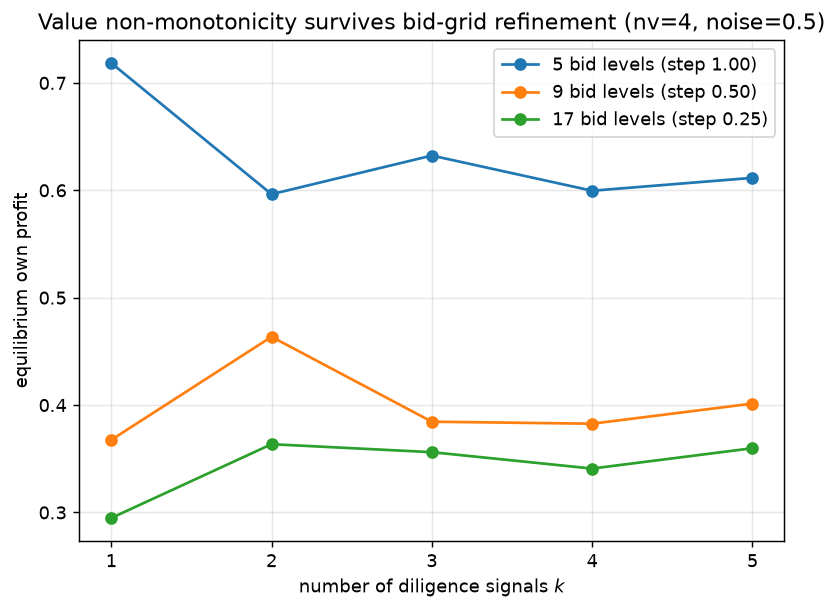}
\caption{The value non-monotonicity is a bid-grid artifact. Re-solving a non-monotone
robustness cell (four values, noise $0.5$) with the bid grid refined from $5$ to $9$ and
$17$ levels over the same bid range: the coarse-grid decline from $k=1$ to $k=2$ reverses
into an increase once the grid is refined, so ``more signals lower profit'' does not
survive refinement and is not an equilibrium property.}
\label{fig:grid-refine}
\end{figure}

\paragraph{Is the cutoff robust?} Because that cutoff is read off one parameterization,
we re-solved it across game sizes (four, five, and six common values) and signal
qualities (noise $0.3$, $0.5$, $0.7$); Fig.~\ref{fig:dilig-robust}. Two things hold and
one does not. \emph{Robust:} the optimal amount of diligence is finite and \emph{falls}
as the per-signal cost rises in every converged parameterization ($7/7$), and acquiring
diligence is valuable on net (equilibrium value at $k=5$ exceeds that at $k=1$ in
$6/7$). \emph{Not robust:} the fine structure. The marginal value of signals is lumpy
but its shape is parameterization-specific: the odd-signal bump that drives the base
game's $k=2$ skip appears in only $3/7$ cases, and in $3/7$ the value is non-monotone in
$k$ (more signals can \emph{lower} a bidder's equilibrium profit at some $k$). We
initially read this non-monotonicity as a common-value equilibrium effect, but it does
not survive refinement of the bid grid: re-solving a non-monotone cell (four values,
noise $0.5$) with the bid grid refined from $5$ to $9$ and $17$ levels over the same bid
range turns the $k{=}1\!\to\!2$ value \emph{decline} ($+0.12$) into an \emph{increase}
(about $-0.07$ to $-0.10$; Fig.~\ref{fig:grid-refine}), so the non-monotonicity is a
coarse-grid quantization artifact, not an equilibrium property (the refined solves carry
more bid actions and settle at NashConv $\sim$$2\times10^{-4}$ rather than $10^{-4}$, but
the reversal is far larger than that residual). We therefore report the specific schedule
(which $k$ a cutoff skips) as a model-internal illustration that is itself sensitive to
the bid grid; the transferable answer is the qualitative one (finite, cost-decreasing
diligence), stated in the model's terms (a stylized constant per-signal cost), not for a
specific transaction.

\begin{figure}[t]
\centering
\includegraphics[width=0.95\linewidth]{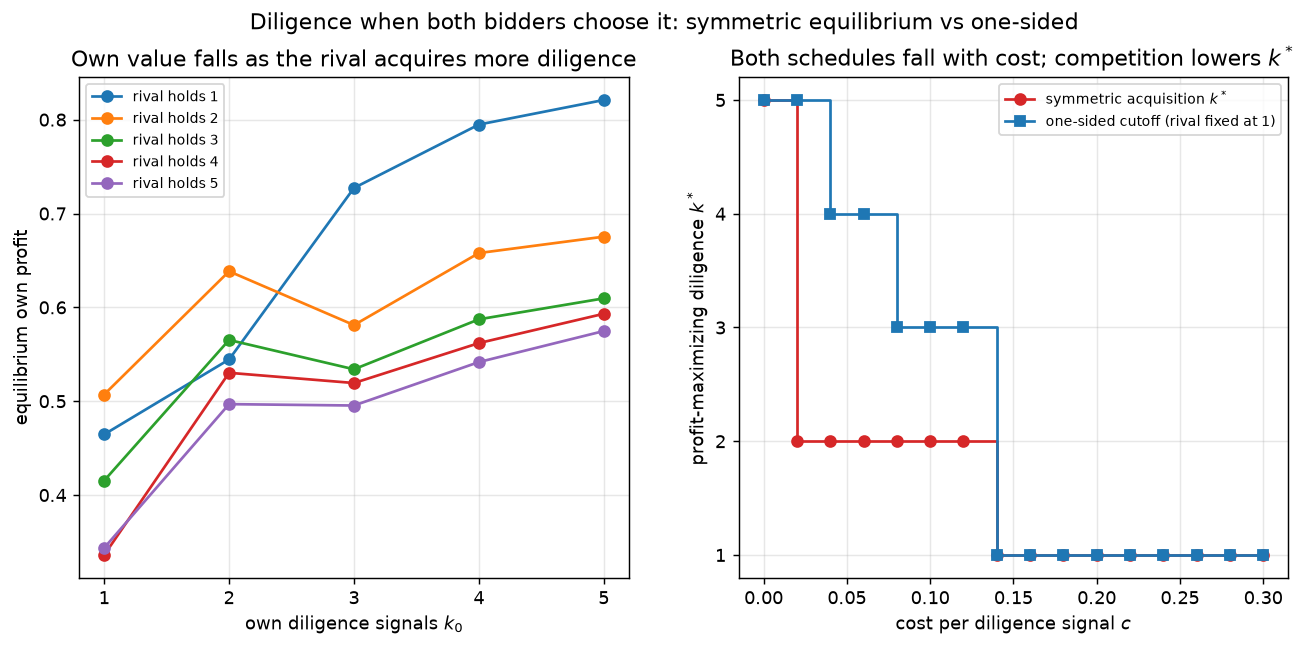}
\caption{Diligence when both bidders choose it. Left: a bidder's equilibrium own profit
rises with its own signals but falls as the rival acquires more, so diligence is partly
rivalrous. Right: the profit-maximizing amount of diligence per signal-cost, in the
symmetric acquisition equilibrium (both bidders choose $k$) versus the one-sided cutoff
(rival fixed at one). Competition makes the symmetric equilibrium more conservative, but
both schedules are finite and fall monotonically with cost.}
\label{fig:acq}
\end{figure}

\paragraph{When both bidders choose diligence.} The curves above fix the rival at one
signal, so they are a one-sided best response, and the toehold result (below) warns that
one-sided analyses overstate effects. We therefore let \emph{both} bidders choose how
much to acquire. Solving the bidding equilibrium for every pair $(k_0, k_1)$ on the base
game gives a value matrix in which a bidder's own profit rises with its own signals but
\emph{falls} as the rival acquires more (own value with five signals drops from
\textbf{0.82} against a one-signal rival to \textbf{0.58} against a five-signal rival),
so diligence is partly rivalrous. At each per-signal cost we then find the symmetric
acquisition equilibrium, the $k^\star$ at which neither bidder gains by deviating when
both acquire $k^\star$ (Fig.~\ref{fig:acq}). The one-sided calculation \emph{overstates}
how much diligence is bought: at intermediate costs the symmetric equilibrium settles at
$k^\star=2$ where the one-sided cutoff reads $3$--$4$, because each extra signal is worth
less when the rival is also well-informed. The title's answer nonetheless survives the
move to a genuine acquisition equilibrium: $k^\star$ is still finite and still falls
monotonically with cost (from $5$ when signals are nearly free, to $2$ at intermediate
cost, to $1$ once $c$ exceeds about $0.14$).

\begin{figure}[t]
\centering
\includegraphics[width=0.6\linewidth]{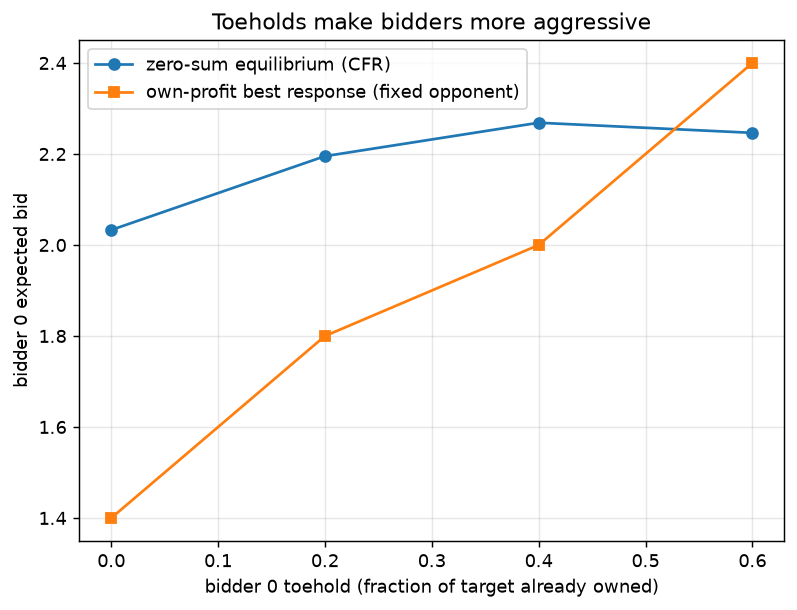}
\caption{Toeholds under the relativized analyses. Both the zero-sum CFR equilibrium bid
and the bid that maximizes bidder~0's \emph{own} profit against a fixed opponent rise
with the toehold $\theta$. At the genuine own-profit equilibrium this aggressiveness is
largely competed away (Fig.~\ref{fig:toe-gs}).}
\label{fig:toe}
\end{figure}

\begin{figure}[t]
\centering
\includegraphics[width=0.95\linewidth]{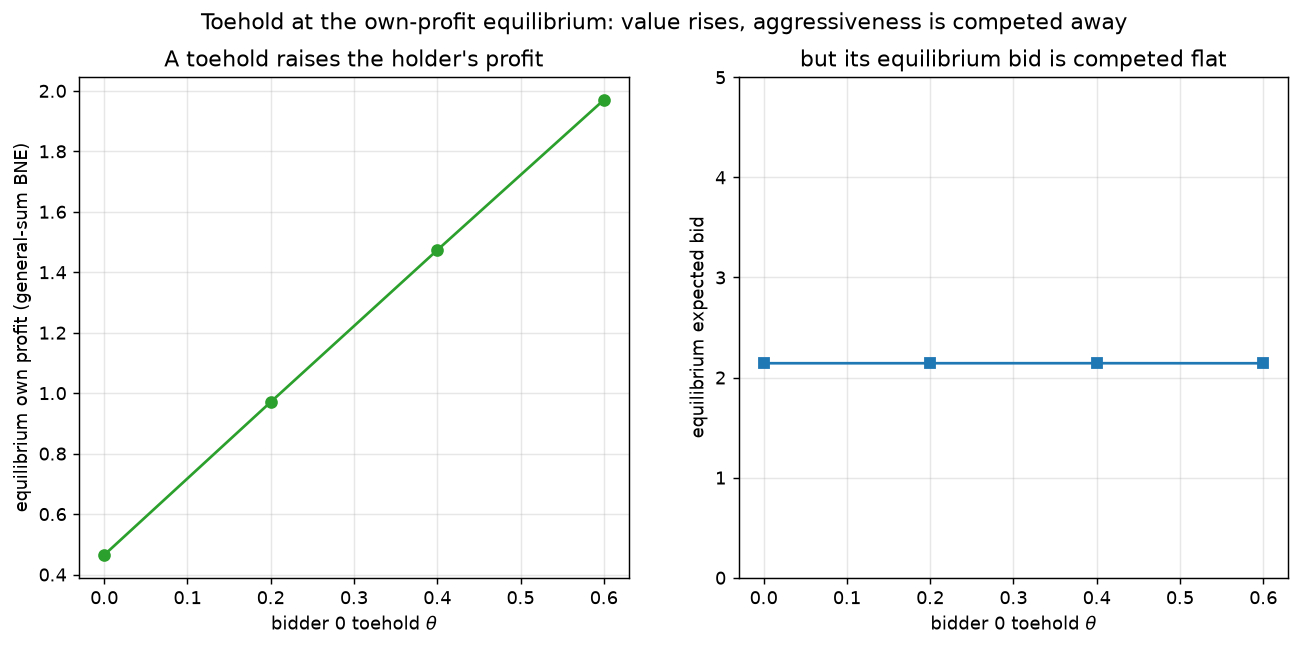}
\caption{The toehold at the genuine own-profit equilibrium. Left: bidder~0's
equilibrium profit rises monotonically with its toehold $\theta$. Right: its
equilibrium expected bid is essentially flat, so the aggressiveness seen under the
relativized analyses (Fig.~\ref{fig:toe}) is competed away once the opponent
re-optimizes. The robust toehold effect is on value, not on equilibrium bidding.}
\label{fig:toe-gs}
\end{figure}

\paragraph{Toeholds.} Bidder~0's expected bid rises with its toehold under the
relativized analyses (Fig.~\ref{fig:toe}): in the zero-sum equilibrium (CFR) it rises
from \textbf{2.03} at $\theta=0$ to \textbf{2.27} at $\theta=0.4$ (then edging down
slightly to \textbf{2.25} at $\theta=0.6$), and its own-profit-maximizing bid against a fixed
opponent rises monotonically from \textbf{1.40} to \textbf{2.40}. At the genuine
own-profit equilibrium, however, the picture is sharper: a toehold raises bidder~0's
equilibrium profit monotonically (from \textbf{0.46} at $\theta=0$ to \textbf{1.97} at
$\theta=0.6$), but its equilibrium \emph{bid} is essentially flat once the opponent
re-optimizes (Fig.~\ref{fig:toe-gs}). This flatness is not an artifact of the coarse bid
grid: refining the grid from $6$ to $20$ levels keeps the equilibrium bid's total
variation across $\theta\in[0,0.6]$ under $0.002$, so it is an equilibrium property. The robust toehold effect is thus on value, not
on equilibrium bidding: the aggressiveness that the zero-sum rendering and a one-sided
best response both show is largely competed away in the symmetric equilibrium. The
aggressiveness channel emphasized by \citet{bulow1999} operates in sequential and
asymmetric contests; in this sealed, simultaneous, symmetric auction it is competed away
at the fixed point, so the flat equilibrium bid is a matter of model scope rather than a
refutation of that channel. That the two objectives disagree on this comparative static
is itself the case for solving the true general-sum equilibrium rather than reading
economics off the zero-sum benchmark.

\paragraph{Scaling: do the learning methods earn their keep?}
\label{sec:scaling}
The headline and cross-game results show the exact solvers winning wherever the game
is tabulatable, so the case for the learning methods must be about \emph{scale}. We
test it directly (Table~\ref{tab:scaling}): on common-value auctions of growing size
we measure the training wall-clock each method needs to drive exploitability to
$0.05$. Exact CFR traverses the whole game tree every iteration, so its time-to-target
grows steeply with the tree, from $2.0$s at $\sim$2k states to $1{,}550$s at
$\sim$365k states (faster than linear: a $\sim$180-fold growth in states costs a
$\sim$780-fold growth in time, because the larger trees also need more iterations to
reach the target, not only more time per iteration). PPO samples a fixed number of
episodes per update regardless of tree size, so its time-to-target is nearly flat
($\sim$190--280s). The cost trends diverge sharply, which is the structural reason to
care about the learning methods. But within the range we can enumerate this does
\emph{not} become a PPO win: CFR is faster on every game we could build, and on the
largest (S12) PPO, with the configuration tuned on the small game and not re-tuned
per scale, fails to reach the target at all (its best exploitability over three seeds
is $0.13$, more than double the $0.05$ target) while CFR reaches it in $1{,}550$s. The
constant per-iteration cost of the learning methods is a real advantage, but it pays
off only past the point where the tree can be enumerated, where CFR cannot run and
exploitability must be estimated approximately. We therefore do not claim the
learning methods beat exact solvers on solvable deal games: we observe \emph{no}
crossover within the enumerable range (the two cost curves diverge but never cross),
and past that range the comparison is no longer like-for-like, because the success
metric must change from exact exploitability to a calibrated lower bound. What we claim
is the weaker, defensible statement: the cost trends make the learning methods the
strongest \emph{available} option once exact methods become infeasible, and exhibiting
that regime (next paragraph) is the natural next step.

\begin{table}[t]
\centering
\caption{Scaling: training wall-clock (seconds) to reach exploitability $0.05$ on
common-value auctions of growing size. Exact exploitability is still computed by tree
enumeration. CFR's cost grows steeply with the tree; PPO's is roughly flat but does
not reach the target on the largest game. PPO seconds are mean $\pm$ std over the
seeds that reached the target (seeds reaching in parentheses); the last column is
PPO's best exploitability reached, mean $\pm$ std over all three seeds. PPO uses the
small-game configuration, not re-tuned per scale.}
\label{tab:scaling}
\small
\begin{tabular}{lrrrr}
\toprule
Game (values, bids) & States & CFR (s) & PPO (s) & PPO best expl. \\
\midrule
S4 (4, 5)   &   2{,}005 &     2.0 & $193 \pm 20$ (3/3) & $0.006 \pm 0.004$ \\
S6 (6, 8)   &  15{,}811 &    31.9 & $238 \pm 48$ (3/3) & $0.022 \pm 0.002$ \\
S8 (8, 10)  &  56{,}905 &   155.7 & $278 \pm 65$ (2/3) & $0.044 \pm 0.010$ \\
S12 (12, 14) & 364{,}765 & 1{,}550.3 & not reached (0/3) & $0.128 \pm 0.043$ \\
\bottomrule
\end{tabular}
\end{table}

\begin{figure}[t]
\centering
\includegraphics[width=0.55\linewidth]{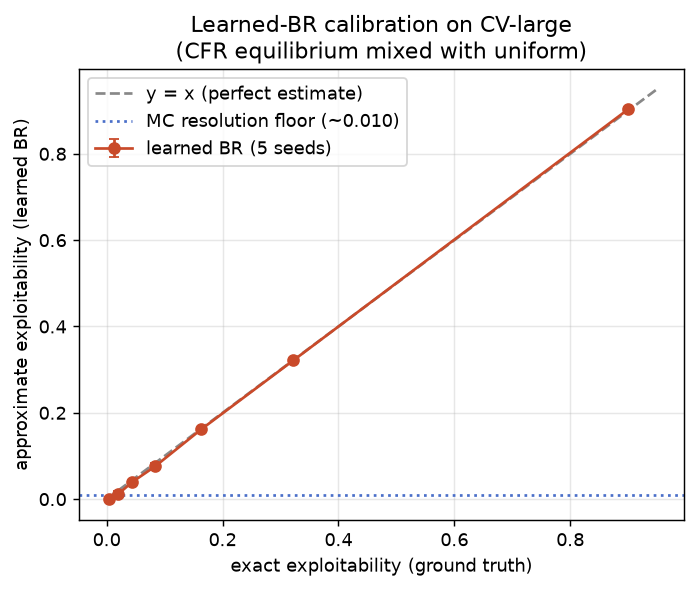}
\caption{Calibrating the learned-best-response estimator on CV-large, where exact
exploitability is available. Policies of known exploitability are built by mixing the
CFR equilibrium with uniform play; markers are the mean over five best-response seeds
and bars the across-seed standard deviation. The estimate tracks the exact value along
the diagonal from $0.043$ ($0.039$) to $0.900$ ($0.903$), staying close down to a
reliable-tracking threshold of about $0.04$ and saturating toward zero below that as it
approaches a $\sim$$0.01$ Monte-Carlo noise floor. This is why a $0.000$ reading on the
intractable game means
exploitability below the resolution floor, not a weak best response.}
\label{fig:calib}
\end{figure}

\paragraph{The intractable regime.} The scaling study stops where the tree can still
be enumerated. To reach the regime the learning methods are actually for, we make the
game genuinely intractable: a \emph{multi-signal} common-value auction in which each
bidder receives $k$ i.i.d.\ noisy signals of $W$ (independent due-diligence
estimates). An information set is the bidder's $k$-tuple of signals, so the number of
information sets per bidder is $\text{values}^{k}$ and the tree is
$\text{values}^{2k+1}\cdot\text{bids}^2$, both exploding with $k$. At $6$ values, $6$
bids, and $k=8$ signals the game has about
$6.1\times10^{14}$ histories and $1.68\times10^{6}$ information sets per bidder: exact
CFR and exact exploitability cannot run, and the strategy cannot be tabulated, but a
network generalizes across signal vectors (the relevant statistic is roughly their
mean). We train PPO and PPG with weight tail-averaging (no tabulation) and measure
exploitability with a \emph{learned best response} \citep{lockhart2019, timbers2022}:
freeze the policy, train a PPO best-responder against it for each player, and Monte-Carlo the
gain from deviating;
half the summed gains estimates exploitability. A learned BR underestimates the true
best response, so this is a lower bound. We calibrate it on CV-large, where exact
exploitability is available as ground truth, by measuring it for policies of \emph{known}
exploitability built by mixing the CFR equilibrium with uniform play
(Fig.~\ref{fig:calib}, five best-response seeds per point). The learned BR tracks the
exact value tightly across more than a decade, from $0.043$ (estimate $0.039 \pm 0.003$)
up to $0.900$ (estimate $0.903 \pm 0.005$), with small across-seed spread throughout, so
the tightness is not single-seed luck. Two distinct scales matter below about $0.04$: a
\emph{Monte-Carlo noise floor} of roughly $0.01$ (the standard error of the estimate at
$20$k evaluation episodes), and a higher \emph{reliable-tracking threshold} of about
$0.04$, below which the estimate begins to under-read before it has reached the noise
floor (at exact $0.020$ it already reads $0.011$). At the CFR equilibrium itself the
estimate is essentially zero ($0.001$ against the exact $0.0037$, below the noise floor).
The estimator is thus tight in the range that matters and saturates only below the
tracking threshold, so a reading of $0.000$ means \emph{exploitability below about
$0.04$}, not a best response too weak to find a deviation.

\begin{figure}[t]
\centering
\includegraphics[width=0.95\linewidth]{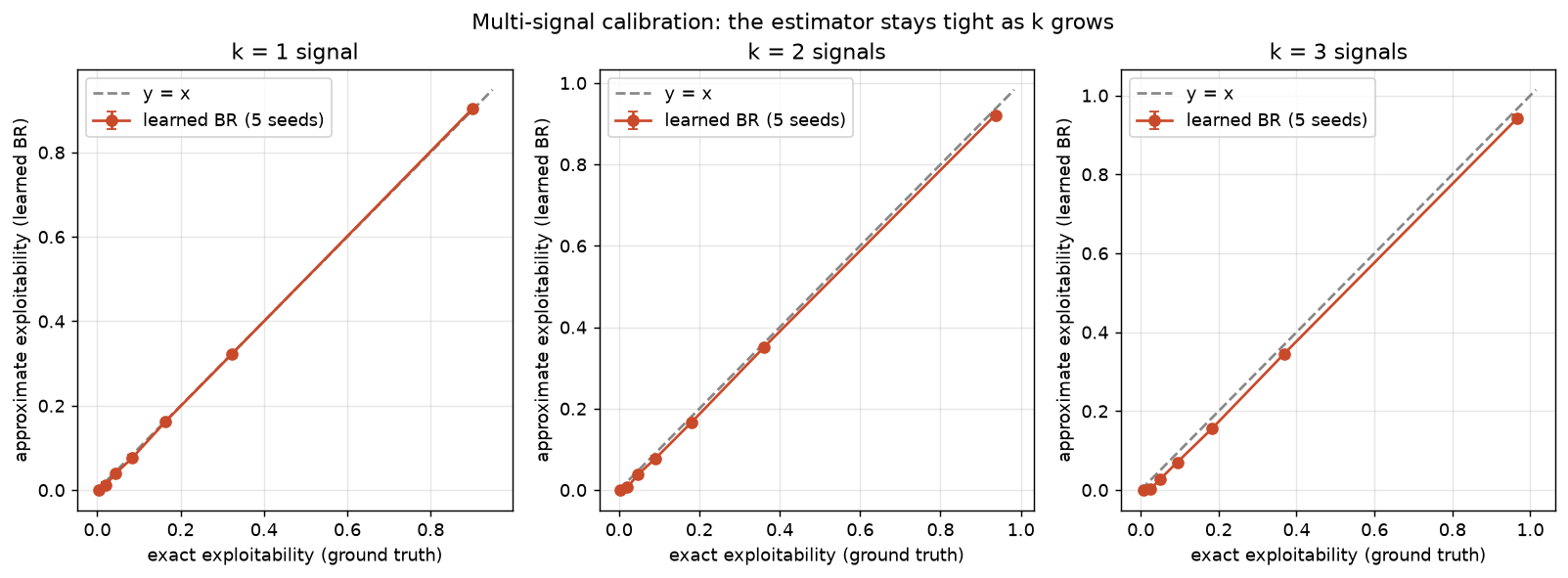}
\caption{Calibrating the same learned-best-response estimator on the
\emph{multi-signal} auction at $k=1,2,3$ signals ($5$ values, $6$ bids), the largest
$k$ still tractable for exact ground truth. As in Fig.~\ref{fig:calib}, policies of
known exploitability are built by mixing the CFR equilibrium with uniform play; markers
are the mean over five best-response seeds and bars are the across-seed standard
deviation. The estimate tracks the exact value along the diagonal across all three
$k$ in the range that matters; the resolution floor rises with $k$ (reliable tracking
from about $0.04$ at $k=1$ to about $0.10$ at $k=3$) but not catastrophically. The
tightness is thus a property of the estimator, not of single-signal structure,
supporting the lower-bound reading on the $k=8$ game.}
\label{fig:calib-multisig}
\end{figure}

\begin{figure}[t]
\centering
\includegraphics[width=0.62\linewidth]{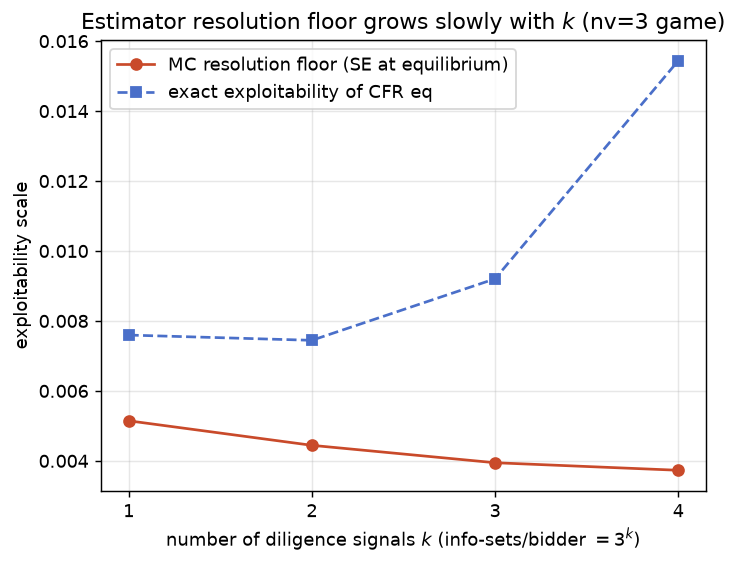}
\caption{Does the resolution floor explode as $k$ grows? On a smaller game ($3$ values,
$4$ bids), we track the floor (the learned-BR estimator's Monte-Carlo standard error at
the near-zero-exploitability CFR equilibrium) over $k=1,\dots,4$. It does not grow with
the information-set count ($3^k$); it edges \emph{down} from about $0.0051$ to $0.0037$,
and the estimator still tracks a known-exploitability mixed policy at every $k$
(Appendix). The floor magnitude is not comparable across game families, but its flat
trend in $k$ is the property the $k=8$ extrapolation on the headline family relies on.}
\label{fig:floor-k}
\end{figure}

One concern remains: CV-large is a single-signal game, whereas the intractable
target carries the $k$-signal information-set structure ($k$-tuples rather than
scalars). To check that the estimator's tightness is a property of the learned BR and
not of single-signal geometry, we repeat the calibration on the multi-signal auction
itself at $k=1,2,3$, the largest $k$ still tractable for exact ground truth ($5$
values, $6$ bids; $k=3$ is already $2.8\times10^{6}$ histories), with five
best-response seeds per mix point (Fig.~\ref{fig:calib-multisig}). The estimate tracks
the exact value closely across all three $k$ in the range that matters, from near $1.0$
down to the low tenths (at $k=3$, exact $0.184$ reads $0.157$ and exact $0.966$ reads
$0.942$). The reliable-tracking threshold does rise with $k$: at $k=1$ the estimate is
still tight at exact $0.05$ ($0.039$), but at $k=3$ the same point reads only $0.028$, so
reliable tracking holds down to about $0.04$ at $k=1$ and about $0.10$ at $k=3$ (the
equilibrium reads $0.000$ against an exact $0.008$ at $k=3$). The calibrated threshold
therefore rises only modestly with $k$ over the range we can check, rather than blowing up, which is the
property the $k=8$ reading relies on: the estimator's tightness survives the
multi-signal information-set structure, not just the single-signal game. The evidence is
strongest at $k\le3$, where we have exact ground truth on the headline family; the $k=8$
reading extrapolates the floor's slow growth past that point, and should be read as such.
As direct support, on a smaller game ($3$ values) solvable to $k=4$ the floor does not
grow in $k$ (it edges down from about $0.0051$ to $0.0037$; Fig.~\ref{fig:floor-k}), so
the floor does not blow up as the information-set count grows.

\begin{figure}[t]
\centering
\includegraphics[width=0.55\linewidth]{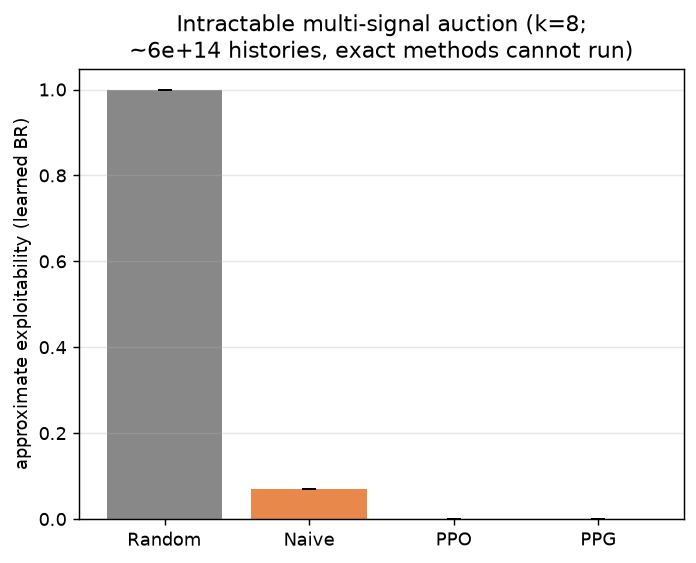}
\caption{Intractable multi-signal common-value auction ($6$ values, $6$ bids, $k=8$
signals; $\sim$$6.1\times10^{14}$ histories, exact methods infeasible). Approximate
(learned-best-response) exploitability of uniform play, a naive unshaded
posterior-mean bidder, and the trained PPO and PPG policies (weight tail-average, mean $\pm$ std over
three seeds). The same BR exploits the references but not the learned policies. Lower
is better.}
\label{fig:intr}
\end{figure}

\paragraph{Result.} On this intractable game, where no exact solver can run, we
compare against two tabulation-free anchors: uniform play and a \emph{naive} bidder
that bids its posterior mean of $W$ without conditioning on the winning event (no
winner's-curse shading). The same
learned BR exploits uniform play almost completely (approximate exploitability
\textbf{0.999}) and the naive bidder partially (\textbf{0.071}, which measures the value
of conditioning on winning that the naive bidder forgoes, one component of the winner's
curse rather than its full magnitude), but
finds no profitable deviation against the trained policies: against both PPO and PPG its
estimate sits at the resolution floor (clamped to \textbf{0.000}) over three seeds
(Fig.~\ref{fig:intr}). The informative comparison is the gap to the anchors: the learned
policies sit far below the naive bidder, which is itself far below uniform play.
Three checks make this credible rather than an artifact of a weak BR: the same BR
clearly exploits both references; doubling its training and widening its network leaves
the anchor estimates unchanged (the BR is at its power ceiling); and, most directly,
the calibrations above show this BR tracks exact exploitability tightly down to about
$0.04$ on the single-signal game and across the multi-signal structure at $k=1,2,3$
(Figs.~\ref{fig:calib},~\ref{fig:calib-multisig}), so a $0.000$ reading places the
learned policies below that resolution and well below the naive bidder's $0.071$, in a
regime exact methods cannot enter. This is the evidence the scaling study set up: once the
game is too large to enumerate, the learning methods still produce a low-exploitability
policy at fixed cost. We report approximate (lower-bound) exploitability and do not
claim a Nash certificate; certifying equilibrium quality here is open.

\begin{table}[t]
\centering
\caption{Final exploitability by method and game (NashConv halved; lower is better).
Exact/tabular solvers (CFR, MMD, PSRO, REINFORCE) are deterministic at fixed budget;
the deep methods (PPO, PPG, DeepPG, DeepCFR, NFSP) are mean $\pm$ std over five
seeds. Deep learners use a fixed episode budget per game ($150$k episodes), smaller
than the $300$k-episode headline deep self-play run; this is why CV-large here
($0.018$--$0.019$) sits slightly above the headline figures ($0.013$--$0.016$) for the
same game. Best learning method per game in bold.}
\label{tab:cross}
\small
\resizebox{\textwidth}{!}{%
\begin{tabular}{lccccccccc}
\toprule
Game & CFR & MMD & PSRO & REINFORCE & PPO & PPG & DeepPG & DeepCFR & NFSP \\
\midrule
CV-small   & 0.0019 & 0.0112 & 0.0027 & 0.0008 & $0.011 \pm 0.010$ & $\mathbf{0.006 \pm 0.009}$ & $0.096 \pm 0.006$ & $0.072 \pm 0.005$ & $0.282 \pm 0.003$ \\
CV-large   & 0.0037 & 0.0236 & 0.0045 & 0.0759 & $0.019 \pm 0.009$ & $\mathbf{0.018 \pm 0.008}$ & $0.235 \pm 0.019$ & $0.122 \pm 0.007$ & $0.508 \pm 0.012$ \\
CV-toehold & 0.0034 & 0.0222 & 0.0075 & 0.3187 & $\mathbf{0.030 \pm 0.011}$ & $0.033 \pm 0.011$ & $0.180 \pm 0.009$ & $0.127 \pm 0.010$ & $0.500 \pm 0.004$ \\
PV         & 0.0019 & 0.0138 & 0.0066 & 0.0603 & $\mathbf{0.014 \pm 0.007}$ & $0.018 \pm 0.012$ & $0.091 \pm 0.003$ & $0.110 \pm 0.003$ & $0.283 \pm 0.006$ \\
\bottomrule
\end{tabular}%
}
\end{table}

\section{Discussion and limitations}

The games are small and stylized abstractions, not calibrated transaction models,
and they are one auction family (common- and private-value first-price), not a broad
suite. The common-value signals are conditionally independent given the asset value
under a specific discretized noise model; affiliated or correlated signal structures
(the general \citealp{milgromweber1982} case) are not explored, and some mechanisms
(for instance that the payoff-relevant statistic is roughly the signal mean) may be
specific to that structure. Our PSRO uses exact best-response oracles rather than RL oracles, and our Deep
CFR stands in for the broader deep-CFR family (ESCHER \citealp{escher2022}, DREAM
\citealp{dream2020}). The zero-sum relative-scoring rendering distorts the underlying
general-sum auction by adding a rivalry term (\S\ref{sec:experiments}); we therefore
read the economic comparative statics off the auction's exact own-profit Bayes-Nash
equilibrium instead (\S\ref{sec:methods-gs}), but the solver benchmark itself still runs
on the zero-sum rendering, where exact two-player exploitability applies, so the
head-to-head method comparison and the economic results are computed on different
objectives. For the intractable regime we report a learned-best-response \emph{lower
bound} on
exploitability, calibrated against exact values where those are available
(\S\ref{sec:experiments}); it is not a Nash certificate, and a certified upper bound
there (for example via a structured or local best response) remains open. The
intractable-regime policy is evaluated through a weight (Polyak) tail-average, which we
validate against the tabular time-average only on CV-large over three seeds
(\S\ref{sec:experiments}); we treat it as a sound stand-in rather than a proven
equivalence at larger scale. Other genuinely intractable instances (continuous
bids/signals or many-round contests) would stress the same estimator further.
Diligence signals are free in the auction games themselves; the cost side enters only
in the diligence-cutoff study (\S\ref{sec:experiments}), where we impose a per-signal
cost and read off the optimal amount, both against a fixed one-signal rival and in a
symmetric acquisition equilibrium where both bidders choose their diligence. That cutoff
uses a stylized constant per-signal cost, not an estimated cost schedule for a particular
deal, so it answers ``how much diligence'' in the model's terms rather than as a
transaction recommendation. Claims about which method is best are empirical statements on these
games, not theorems.

\section{Conclusion}

On zero-sum auction games from the takeover literature, the generic policy-gradient
methods PPO and PPG are the strongest learning solvers, clearly beating the deep
fictitious-play and deep-CFR baselines on commodity CPU with no frontier-model spend;
this gap survives scoring every method without the policy-gradient family's burn-in, and
the one tuning asymmetry (we swept the policy-gradient learning rate, not the baselines')
is stated plainly. Solving the auction's own-profit Bayes-Nash equilibrium exactly, we
quantify the value of due diligence and the cost at which acquiring more of it stops paying. The
honest qualification is that on every game
small enough to tabulate the exact solvers (CFR, MMD, PSRO) are both more accurate and
faster: the policy-gradient methods win the learning bracket, not the overall race. A
scaling study shows why they may still matter, their per-target cost staying flat
while exact CFR's grows steeply with the game tree, and we take the first step into
the regime that settles the question: on a multi-signal auction too large to
enumerate, where exact solvers cannot run, PPO and PPG drive a learned-best-response
exploitability estimate to its resolution floor, below a naive unshaded bidder, with
the estimator calibrated against exact values on the games where both can be computed.
Certifying equilibrium quality there, beyond a lower bound, is the open problem; the
shared abstraction is a foundation for these richer games.

\appendix
\section{Hyperparameters}
\label{app:hparams}

All deep methods use a single hidden layer of $64$ units and the Adam optimizer,
except Deep CFR, which uses two $(64, 64)$ networks (its advantage and policy nets);
full solver settings are in Table~\ref{tab:hparams}. Means are over five seeds
$\{0,1,2,3,4\}$ for the headline and cross-game studies and three seeds $\{0,1,2\}$ for
the scaling, intractable, and weight-averaging studies. Both the single-signal calibration
(Fig.~\ref{fig:calib}) and the multi-signal calibration (Fig.~\ref{fig:calib-multisig})
use five best-response seeds per mix point with across-seed standard-deviation bars. All
tail-averages discard the first $50\%$ of training as burn-in.

\paragraph{Episode budgets.} Headline deep self-play $300$k; cross-game table $150$k;
scaling-study PPO $600$k (time-to-target is reached well before the cap);
intractable-regime PPO/PPG $300$k.

\paragraph{General-sum equilibrium solver.} The economic comparative statics are read
off the auction's own-profit Bayes-Nash equilibrium (\S\ref{sec:methods-gs}), found by
fictitious play on the factorized tensor model (linear averaging of pure best responses,
up to $60$k iterations with an early stop at NashConv $<10^{-4}$). The
information-asymmetry study sweeps bidder~0's signal noise over $\{0.0, 0.25, 0.5, 0.75, 1.0\}$ with
bidder~1 fixed at $0.5$; the toehold study sweeps $\theta$ over $\{0.0, 0.2, 0.4, 0.6\}$;
the diligence study sweeps bidder~0's number of signals over $\{1,2,3,4,5\}$ (each of
noise $0.5$) with the rival fixed at one signal, and the cost cutoff is taken over a
per-signal cost grid in steps of $0.02$ up to $0.30$. All three use $5$ values and $6$
bids. The achieved NashConv is $\sim$$10^{-4}$ except at the fully-informed corner
(noise $0$), where it is $\sim$$8\times10^{-3}$. The diligence-cutoff robustness sweep
(Fig.~\ref{fig:dilig-robust}) repeats the diligence study over $\text{values}\in\{4,5,6\}$
(with $\text{bids}=\text{values}+1$) and $\text{noise}\in\{0.3,0.5,0.7\}$, allowing up to
$400$k fictitious-play iterations; $2$ of the $9$ cells (both at noise $0.3$) did not
reach tolerance and are excluded from the robustness counts. To firm up the $k=8$
extrapolation of the estimator's resolution floor, which the calibration only reaches at
$k\le3$ on the headline family, we additionally track the floor over $k=1,\dots,4$ on a
smaller game ($3$ values, $4$ bids), with $3$ seeds (Fig.~\ref{fig:floor-k}); $k=4$ here
($\sim$$3.1\times10^{5}$ histories) extends a full step past the headline calibration,
and $k=5$ ($\sim$$2.8\times10^{6}$ histories), though enumerable in principle, is already
impractically slow for exact CFR, itself an instance of the scaling wall. The zero-sum toehold bids in
Fig.~\ref{fig:toe} are the CFR equilibrium ($600$ iterations) and the own-profit best
response against the $\theta{=}0$ CFR equilibrium opponent, retained for comparison.
The symmetric acquisition equilibrium (Fig.~\ref{fig:acq}) solves the bidding
equilibrium for every signal pair $(k_0, k_1)$ with $k_0, k_1\in\{1,\dots,5\}$ on the
base game and, at each per-signal cost, reports the symmetric pure-strategy acquisition
equilibria $k^\star$ (no unilateral deviation in $k$ improves net payoff). The
grid-refinement check (Fig.~\ref{fig:grid-refine}) re-solves a non-monotone robustness
cell ($4$ values, noise $0.5$) with the bid grid set to $5$, $9$, and $17$ levels evenly
spaced over the same bid range $[0,\text{values}]$; the larger action sets converge more
slowly (NashConv $\sim$$2\times10^{-4}$).

\paragraph{Software and threading.} All experiments run in a pinned Docker image:
Python $3.12.13$, CPU-only PyTorch $2.12.1$, OpenSpiel $1.6.15$, NumPy $<2.0$. To make
the wall-clock comparisons reproducible we pin threading to eight cores
($\texttt{OMP\_NUM\_THREADS}=\texttt{MKL\_NUM\_THREADS}=8$ in the image, and the deep
benchmark script additionally calls $\texttt{torch.set\_num\_threads}(8)$), so
floating-point reduction order is fixed across runs.

\begin{table}[h]
\centering
\small
\begin{tabular}{ll}
\toprule
Method & Settings \\
\midrule
PPO & lr $3\times10^{-3}$, clip $0.2$, $4$ epochs, $4$ minibatches, entropy $0.01$, value
coef $0.5$, grad clip $0.5$, batch $256$ episodes \\
PPG & as PPO, plus $n_{\text{policy}}{=}8$ policy phases per auxiliary phase, $6$
auxiliary epochs, behavioral-cloning coef $1.0$ \\
DeepPG & OpenSpiel PolicyGradient, RPG loss, $\pi$-lr $5\times10^{-3}$, critic-lr
$5\times10^{-2}$, batch $16$ \\
REINFORCE & tabular, lr $0.5$, entropy $0.01$, batch $256$ episodes, $200$ iterations
(cross-game) / $300$ (convergence figure) \\
Deep CFR & lr $10^{-3}$, $80$ iterations, $100$ traversals, $600$ advantage / $600$
policy train steps, advantage and strategy batch $2048$ \\
NFSP & rl-lr $10^{-2}$, sl-lr $10^{-2}$, anticipatory $0.1$, batch $128$, reservoir
$2\times10^{6}$, replay $2\times10^{5}$, $\varepsilon$ $0.6{\to}0.01$ over the horizon \\
CFR & $400$ iterations (headline, cross-game, calibration); $600$ for the zero-sum
toehold comparison; $300$ for the tabular-convergence figure; up to $1{,}000$ in the
scaling time-to-target study \\
MMD & entropy-regularized, stepsize $1.0$, temperature $\alpha{=}0$ (targets Nash) \\
PSRO & $15$ iterations, exact best-response oracle, projected-replicator-dynamics meta-solver \\
\bottomrule
\end{tabular}
\caption{Solver hyperparameters. The policy-gradient family (PPO, PPG, DeepPG) shares a
swept learning rate; the other deep baselines use author-recommended settings adapted
to the short CPU budget (\S\ref{sec:experiments}).}
\label{tab:hparams}
\end{table}

\paragraph{Intractable-regime estimator.} The learned best response trains for $250$
update batches of $256$ episodes each (same PPO settings, $64$-wide); approximate
exploitability is Monte-Carlo'd over $20$k evaluation episodes; the evaluated policy is
the weight tail-average. The calibration study (Fig.~\ref{fig:calib}) uses the same
estimator settings, with the reference policies built from a $400$-iteration CFR
equilibrium mixed with uniform play. The multi-signal calibration
(Fig.~\ref{fig:calib-multisig}) uses the same estimator and mix grid on the $5$-value,
$6$-bid auction at $k=1,2,3$ signals (up to $2.8\times10^{6}$ histories at $k=3$), with
the reference CFR run for $400$ iterations at $k{=}1,2$ and $200$ at $k=3$, and five
best-response seeds per mix point.

\bibliographystyle{plainnat}
\bibliography{references}

\end{document}